%% file: main.tex
\documentclass[runningheads]{llncs}
\usepackage{graphicx}
\usepackage[flushleft]{threeparttable}
\usepackage{multirow}
\usepackage{makecell}
\usepackage{amssymb}
\usepackage{amsmath}
\usepackage{xcolor} 
\usepackage{marvosym}

\definecolor{linkcolor}{HTML}{ED1C24}
\definecolor{pearDark}{HTML}{2980B9}

\usepackage[pagebackref=false,breaklinks=true,colorlinks,urlcolor=blue,citecolor=blue,linkcolor=linkcolor,bookmarks=false, ]{hyperref}
\usepackage{color}
\usepackage{colortbl}

\usepackage{pifont}
\newcommand{\cmark}{\ding{51}}
\newcommand{\xmark}{\ding{55}}

\newcommand{\rui}[1]{{\color[rgb]{0.0,0,0}#1}}

\begin{document}
\title{SR-Mamba: Effective Surgical Phase Recognition with State Space Model}
%
\titlerunning{Effective Surgical Phase Recognition with State Space Model}

%
\author{Rui Cao\inst{1,2} \and
Jiangliu Wang\inst{1,2} \and
Yun-Hui Liu\inst{1,2}\textsuperscript{(\Letter)}}

\authorrunning{R. Cao et al.}

\institute{Department of Mechanical and Automation Engineering, \\
The Chinese University of Hong Kong, Hong Kong, China \\
\and
T Stone Robotics Institute, The Chinese University of Hong Kong, \\ Hong Kong, China
\\ \email{yhliu@mae.cuhk.edu.hk}}


%
\maketitle              
\begin{abstract}
Surgical phase recognition is crucial for enhancing the efficiency and safety of computer-assisted interventions. One of the fundamental challenges involves modeling the long-distance temporal relationships present in surgical videos.
Inspired by the recent success of Mamba, a state space model with linear scalability in sequence length, this paper presents \textbf{SR-Mamba}, a novel attention-free model specifically tailored to meet the challenges of surgical phase recognition.
\rui{In SR-Mamba, we leverage a bidirectional Mamba decoder to effectively model the temporal context in overlong sequences.}
Moreover, the efficient optimization of the proposed Mamba decoder facilitates single-step neural network training, eliminating the need for separate training steps as in previous works.
This single-step training approach not only simplifies the training process but also ensures higher accuracy, even with a lighter spatial feature extractor. 
Our SR-Mamba establishes a new benchmark in surgical video analysis by demonstrating state-of-the-art performance on the Cholec80 and CATARACTS Challenge datasets. 
The code is accessible at \url{https://github.com/rcao-hk/SR-Mamba}.

\end{abstract}

\section{Introduction}
Intelligent computer-assisted intervention (CAI) systems have significantly improved patient care within contemporary operating theaters by augmenting safety and enhancing the quality of procedures. At the heart of CAI's functionality is the ability to recognize surgical activities undertaken by surgeons, an aspect critical for the system's effectiveness. Surgical workflow recognition, focusing on identifying different phases and actions during surgery, is pivotal~\cite{maier2022surgical,demir2023deep,vercauteren2019cai4cai}. It not only facilitates the development of surgical expertise but also boosts procedural efficiency and safety, offering real-time feedback to the surgical team.

Temporal relations are crucial to analyze surgical workflows.
Currently, due to the long interdependencies of surgical phases, methods for surgical phase recognition often rely on long sequences of frame input. 
However, the duration of surgical videos typically spans several hours, which poses a challenge for training models in an end-to-end fashion. 
Traditionally, this challenge has been addressed through a two-step approach~\cite{demir2023deep}. 
Initially, a spatial feature extractor, often employing ResNet50~\cite{he2016deep}, is trained to encode spatial features of the surgical scene. Subsequent temporal feature extraction is then performed using the encoded features. 
Temporal modeling techniques generally fall into three categories: Recurrent Neural Networks (RNNs), Temporal Convolution Networks (TCNs), and Transformers.

\begin{figure}[t]
\includegraphics[width=\textwidth]{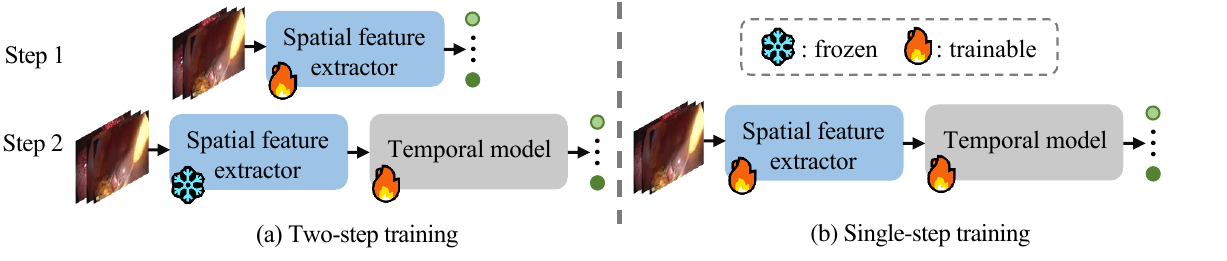}
\caption{(a) Previous two-step \rui{training strategy}. (b) Our proposed single-step \rui{method} jointly trains a spatial feature extractor and a temporal model in a single step.} \label{fig:framework}
\end{figure}

Early efforts in this domain utilized Long Short-Term Memory (LSTM) networks, establishing a two-step CNN-LSTM architecture as a standard. Following this, Czempiel et al.~\cite{czempiel2020tecno} replaced LSTM with TCN, achieving a 2$\%$ improvement in accuracy. Ramesh et al.~\cite{ramesh2021multi} introduced a multi-stage TCN that further enhanced performance. More recently, the advent of self-attention mechanisms~\cite{vaswani2017attention} has led to the adoption of Transformer, which replaced LSTM in the work of Czempiel et al.~\cite{czempiel2021opera}, yielding a 1$\sim$2$\%$ accuracy increase over CNN-TCN models. Gao et al.~\cite{gao2021trans} introduced a TCN Transformer network that merges spatial embeddings from ResNet with temporal embeddings from TCN for improved analysis. Meanwhile, Liu et al.~\cite{liu2023lovit} substituted ResNet with a Vision Transformer~\cite{dosovitskiy2020vit}, setting a new benchmark for state-of-the-art performance in the field.

Recently, state space models (SSMs) have garnered considerable attention for their proficiency in modeling long-range dependencies. This is attributed to their convolutional computation approach, which ensures near-linear computational efficiency~\cite{gu2020improving}. A noteworthy development in this area is Mamba~\cite{mamba}, which integrates time-varying parameters into the SSMs and introduces a hardware-aware algorithm to facilitate efficient training and inference processes. This selection mechanism boosts the performance of SSMs and suggests its potential as a promising alternative to Transformer. 

In this paper, we propose SR-Mamba, a novel approach leveraging SSMs for efficient surgical phase recognition. We first present a bidirectional Mamba decoder for advanced temporal modeling. Subsequently, building upon the favorable features of Mamba decoder, illustrated in Fig.~\ref{fig:framework} (b), we develop a single-step \rui{training method} that jointly trains the spatial feature extractor and temporal model in an end-to-end fashion. Our experimental results show that the proposed model achieves superior performance, even with a simpler training procedure and a lighter spatial feature extractor, highlighting the bidirectional Mamba decoder's exceptional temporal modeling prowess. 

\section{Methodology}
\begin{figure}[t]
\includegraphics[width=\textwidth]{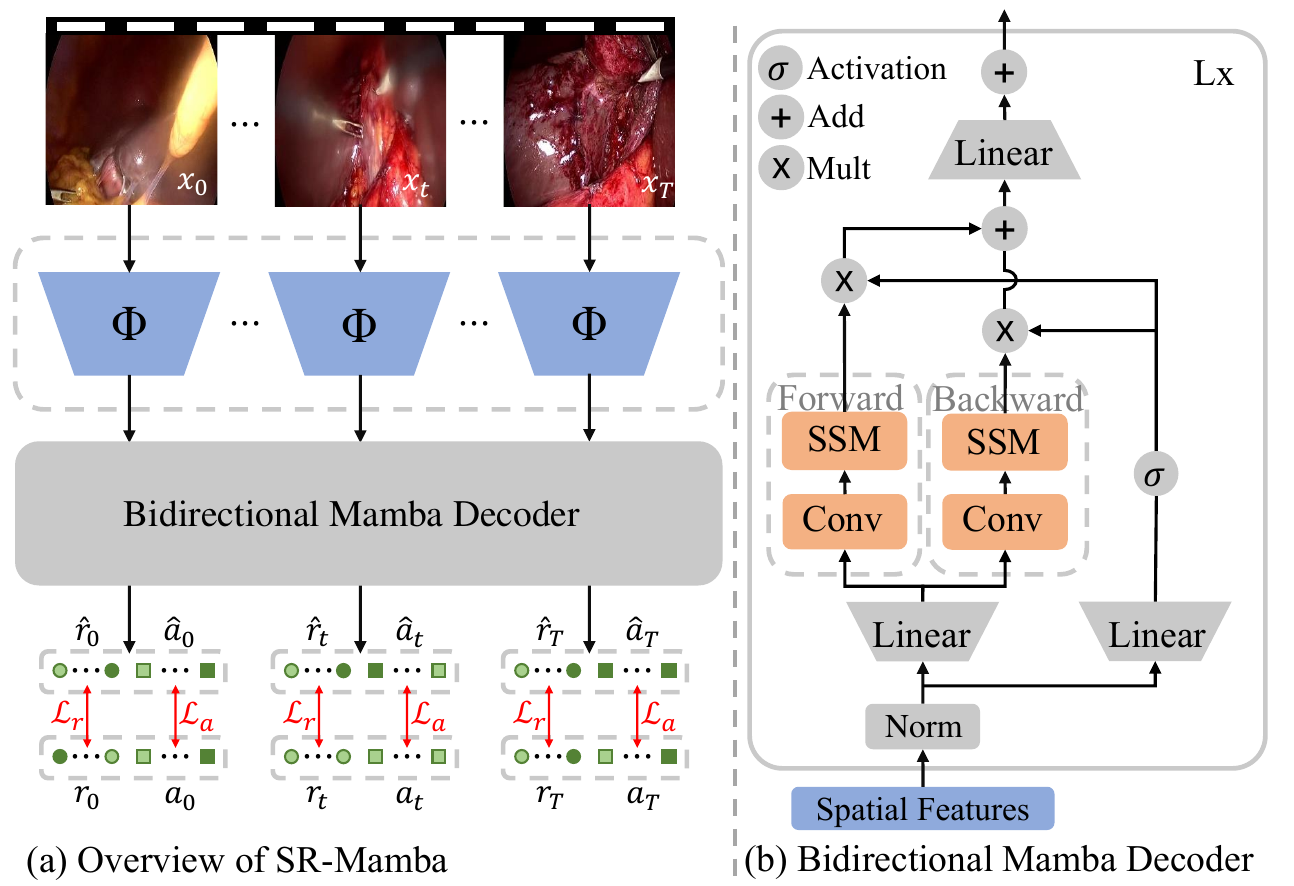}
\caption{Overview of the proposed SR-Mamba architecture for surgical phase recognition. (a) Sequential processing of surgical video frames $\mathbf{X}=\{x_{t}\}_{t=1}^{T}$ through a spatial feature extractor $\mathbf{\Phi}$. The resulting embeddings are then fed into a Bidirectional Mamba Decoder. (b) Detailed view of the Bidirectional Mamba Decoder.} \label{fig:overview}
\end{figure}

In this section, we outline our proposed SR-Mamba. Initially, we provide background on the SSMs. We then introduce our model with single-step training, which employs a bidirectional SSM for temporal modeling. Fig.~\ref{fig:overview} illustrates the framework of our proposed SR-Mamba, with further details presented in the following subsections.

\subsection{Preliminaries: State Space Models}
Drawing from the principles of the SSM, as formulated in Eq.~\ref{eq:lti}, Mamba~\cite{mamba} utilizes a linear Ordinary Differential Equation (ODE) to map an input sequence of length $T$ as: $x(t) \in \mathbb{R}^{T} \mapsto y(t) \in \mathbb{R}^{T}$ by a hidden state $h(t) \in \mathbb{R}^N$.
\begin{equation}
\begin{aligned}
\label{eq:lti}
h'(t) &= \mathbf{A}h(t) + \mathbf{B}x(t), \\
y(t) &= \mathbf{C}h(t),
\end{aligned}
\end{equation}
where $\mathbf{A} \in \mathbb{R}^{N \times N}$, $\mathbf{B} \in \mathbb{R}^{N \times 1}$, and $\mathbf{C} \in \mathbb{R}^{1 \times N}$ denote the state matrix, input matrix, and output matrix, respectively. For practical computation in discrete settings, Mamba employs the zero-order hold (ZOH) technique, rewriting the Eq.~\ref{eq:lti} as:

\begin{equation}
\begin{aligned}
\label{eq:discrete_lti}
h_t &= \mathbf{\overline{A}}h_{t-1} + \mathbf{\overline{B}}x_{t}, \\
y_t &= \mathbf{C}h_t.
\end{aligned}
\end{equation}

The parameters for the discretization process are computed as shown in Eq.~\ref{eq:zoh}, allowing the discrete system to capture the dynamics of its continuous counterpart over discrete time intervals $\Delta$.

\begin{align}
\label{eq:zoh}
&\mathbf{\overline{A}} = \exp{(\mathbf{A}\Delta)},
&\mathbf{\overline{B}} = (\mathbf{A}\Delta)^{-1}(\exp{(\mathbf{A}\Delta)} - \mathbf{I}) \mathbf{B}\Delta.
\end{align}

In the discrete formulation, Mamba models the matrix $\mathbf{\overline{A}}$ as the evolution parameter and $\mathbf{\overline{B}}$ and $\mathbf{C}$ as the projection parameters. The output $ \mathbf{y} $ is then computed using a structured convolutional kernel $ \overline{\mathbf{K}}$ and given $T$ length input sequence $\mathbf{x}$, as detailed in Eq.~\ref{eq:conv}.

\begin{align}
\label{eq:conv}
&\mathbf{\overline{K}} = (\mathbf{C}\mathbf{\overline{B}}, \mathbf{C}\mathbf{\overline{A}}\mathbf{\overline{B}}, \dots, \mathbf{C}\mathbf{\overline{A}}^{T-1}\mathbf{\overline{B}}),
&\mathbf{y} = \mathbf{x} * \mathbf{\overline{K}}.
\end{align}

To accommodate time-variant dynamics and enhance model capacity, matrices such as $\Delta$, $\mathbf{B}$, and $\mathbf{C}$ are defined as functions of the input. Moreover, Mamba employs a modern-hardware-friendly parallel scan algorithm~\cite{smith2023simplified} to circumvent the sequential recurrence typically associated with time-varying modeling. These strategic design choices significantly empower selective long-sequence modeling and greatly facilitate efficient GPU optimization efforts.

\subsection{SR-Mamba Model}
\subsubsection{\rui{Single-Step Training.}}
Motivated by the goal of circumventing the complexities associated with two-step training and the fine-tuning of related parameters, we propose a single-step \rui{training method} for surgical video recognition. Instead of adopting a multi-step approach, \rui{we directly input the RGB images of a video sequence} for simultaneous training of both the spatial feature extractor and the temporal model. This method not only simplifies the training \rui{process} but also capitalizes on the full potential of Mamba's linear scalability and efficient training capabilities.

For a surgical video sequence of length $T$, represented as $\mathbf{X}=\{x_{t}\}_{t=1}^{T} \in \mathbb{R}^{T\times H\times W\times C}$ where $t$ denotes the frame index, we derive spatial embeddings $\mathbf{F}=\{f_{t}\}_{t=1}^{T}$ using the feature extractor $\mathbf{\Phi}$:$\mathbf{F}=\mathbf{\Phi}(\mathbf{X})$. Opting for a lightweight ResNet34~\cite{he2016deep} as our $\mathbf{\Phi}$, instead of the more traditional ResNet50 or Vision Transformer (ViT)~\cite{dosovitskiy2020vit} employed in recent works~\cite{gao2021trans,demir2023deep,liu2023lovit}, allows us to achieve a balance between maintaining performance and managing the computational demands inherent in single-step training. Consequently, these feature embeddings are formalized as $\mathbf{F} \in \mathbb{R}^{T \times 512}$, reflecting the chosen ResNet's output dimensions.

Then the $\mathbf{F}$ for \rui{the video sequence} is fed into the bidirectional Mamba decoder $\mathbf{\Psi}$, shown as Fig.~\ref{fig:overview} (a). In acknowledging the synergistic relationship between recognition and anticipation in the context of surgical phase recognition, we introduce anticipation prediction as an auxiliary task to augment performance. To obtain the ground truth for anticipation, we adopt the approach from~\cite{yuan2021surgical,jin2022trans}, reformulating it to predict the remaining time. Consequently, our model generates two distinct types of prediction outputs: workflow recognition, represented as \( \mathbf{R}=\{r_{t}\}_{t=1}^{T} \), and workflow anticipation, represented as \( \mathbf{A}=\{a_{t}\}_{t=1}^{T} \). These are formalized as follows:
\begin{equation}
\label{eq:predicton}
\mathbf{\hat{R}}, \mathbf{\hat{A}} = \mathbf{\Psi}(\mathbf{F}).
\end{equation}

\subsubsection{Bidirectional Mamba Decoder.} Inspired by~\cite{vim,liu2024vmamba}, our decoder employs bidirectional Mamba layers to capture the temporal relation in surgical videos, as illustrated in Fig.~\ref{fig:overview} (b). It begins by normalizing the sequential input $\mathbf{x}$ and then processing the $\mathbf{x}$ from the forward and backward directions. In each directional path, a 1-D convolution is applied to $\mathbf{x}$, producing $\mathbf{x}'_{i}$, with $i$ serving as an identifier for either the forward or backward direction. Subsequently, $\mathbf{x}'_{i}$ undergoes linear projection to get $\mathbf{B}_{i}$, $\mathbf{C}_{i}$, and $\mathbf{\Delta}_{i}$. The outcomes, $\mathbf{y}_{forward}$ and $\mathbf{y}_{backward}$, are gated by $\mathbf{z}'$ (generated by activating $\mathbf{z}$ via SiLU~\cite{hendrycks2023gaussian}) and merged to produce the final output sequence $\mathbf{y}_{\mathtt{l}}$ at layer $\mathtt{l}$. For the sake of simplicity and clarity, we will omit the subscript $\mathtt{l}$ from all further discussions.

\subsection{Optimization}
We use the cross-entropy loss for the workflow recognition task and SmoothL1 loss for \rui{the workflow anticipation as an auxiliary task}, as defined in Eq.~\ref{eq:loss}. \rui{See Sec.~\ref{sec:ablation} for the effect of incorporating anticipation loss.}
The overall loss is computed as $\mathcal{L} = \lambda_1\mathcal{L}_r + \lambda_2\mathcal{L}_a$, with $\lambda_1$ and $\lambda_2$ as hyperparameters to balance the contributions of each loss component.
\begin{align}
\label{eq:loss}
&\mathcal{L}_r =-\sum_{t=1}^{T}r_{t} \log(\hat{r_{t}}), 
&\mathcal{L}_a =-\sum_{t=1}^{T}\mathrm{SmoothL1}(\hat{a_{t}}, a_{t}).
\end{align}

For sequences longer than max length \(N_{\max}\), we adopt a sampling method that retains keyframes: each phase transition (\(x_s\)) and its adjacent frame (\rui{\(x_{s-1}\)}), where $s$ stands the transition frame's time. For the intervals between transitions, we uniformly sample additional frames, maintaining the phase representation proportional to its duration in the video. This strategy effectively balances the sequence's coverage while managing computational load.




\section{Experiments}
\subsection{Implementation Details}
\subsubsection{Datasets.} 
We evaluate our model on two benchmark datasets for surgical phase recognition: Cholec80~\cite{cholec80} and the CATARACTS~\cite{cataracts}. The Cholec80 dataset consists of 80 endoscopic videos of cholecystectomy procedures with annotated 7 surgical phases. For our experiments, we follow the most commonly used train-test split of 40 videos for training and the remaining 40 for testing~\cite{czempiel2020tecno,gao2021trans,jin2022trans}. The videos are recorded at a resolution of $1920\times1080$ and are downsampled to a frame rate of 1 FPS to reduce computational cost. CATARACTS comprises 50 microscopic videos with $1920\times1080$ resolution of cataract surgery, annotated with more fine-grained steps: 18 surgical steps and 1 idle step. Following the challenge~\cite {cataracts}, We use 25 videos for training, 5 for validating, and 20 for testing.



\subsubsection{Training.}
Our model is implemented in PyTorch and all training and evaluations are conducted on a single NVIDIA RTX 3090 GPU. For preprocessing, frames from both datasets are resized to $250\times250$ pixels. We apply data augmentation techniques such as $224\times224$ cropping, random mirroring, and color jittering. We employ ResNet34, pre-trained on the ImageNet-1k dataset~\cite{imagenet15russakovsky}, as the backbone for feature extraction, which is facilitated by the TIMM library~\cite{rw2019timm}. To make memory usage manageable, we freeze the first two layers of ResNet34, and keep the remaining layers trainable. To enhance the model's generalization capabilities and mitigate overfitting, we integrate Stochastic Depth~\cite{huang2016deep} with a drop path rate of 0.1. Our single-step \rui{training} undergoes end-to-end training utilizing the AdamW optimizer~\cite{loshchilov2017decoupled} coupled with a step learning rate scheduler that reduces the learning rate from 2e-4 by half every 50 epochs. We set the maximum sequence length, $N_{\max}$, to 2048, $\lambda_1 = 0.5$, $\lambda_2 = 1$. For the bidirectional Mamba decoder, we employ 2 layers for the Cholec80 and 1 layer for CATARACTS.

\begin{table}[t]
	\centering\footnotesize
        \setlength{\tabcolsep}{4pt}
	\caption{The results ($\%$) of different methods on Cholec80 dataset~\cite{cholec80}.}
        \scalebox{0.95}{
        \input{Tables/cholec80_results}
        }
\label{tab:cholec80_results}
\end{table}

\subsubsection{Evaluation Metrics.} 
Consistent with prior studies~\cite{czempiel2020tecno,gao2021trans,jin2022trans}, we employ four metrics to evaluate our model on the Cholec80 dataset: accuracy (AC), precision (PR), recall (RE), and the Jaccard index (JA). For the CATARACTS dataset, we adopt the F1 score in line with the CATARACTS2020 challenge's evaluation~\cite{cataracts}. \rui{Additionally, we report the parameter counts of our and other SOTAs for a detailed comparison of model complexities.}

\begin{table}[ht]
	\centering\footnotesize
        \setlength{\tabcolsep}{4pt}
	\caption{Average F1 score of different methods on CATARACTS challenge dataset~\cite{cataracts}.}
        \scalebox{0.94}{
        \input{Tables/cata_results}
        }
        \begin{tablenotes}
          \small
          \item \textit{Note: the $\ast$ denotes the methods using extra labels.}
        \end{tablenotes}
\label{tab:cata_results}
\end{table}

\subsection{Comparison with State-of-the-Arts}
We report the quantitative results in Tab.~\ref{tab:cholec80_results}, and Tab.~\ref{tab:cata_results} for Cholec80 and CATARACTS respectively. Tab.~\ref{tab:cholec80_results} shows the representative methods for Cholec80. Our SR-Mamba outperforms all listed models, achieving an AC of 92.6\% ± 8.6 and a JA of 81.5\% ± 8.6, \rui{while utilizing fewer parameters.} Notably, it surpasses MTRCNet-CL~\cite{jin2020multi} by 3.4 percentage points in AC, even without the use of additional tool labels that MTRCNet-CL relies on. Impressively, our model achieves these results using a lighter ResNet34 with a frozen first two layers, compared to the ResNet50 used by most competitors, proving that advanced sequence modeling can compensate for a less complex spatial feature extractor. Moreover, our method exceeds Not End-to-End~\cite{yi2022not} by 4.3 percentage points in JA, while employing a more streamlined pipeline without the need for a refinement stage. 

\begin{figure}[t]
\includegraphics[width=\textwidth]{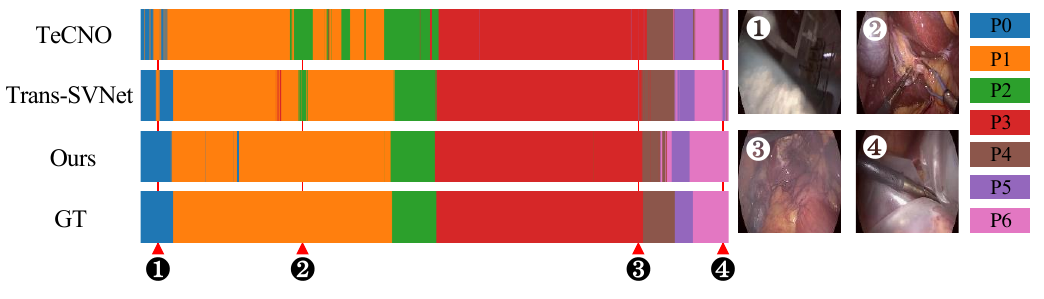}
\caption{Color-coded ribbon illustration of one complete surgical video from Cholec80 dataset~\cite{cholec80}. The time axes have been adjusted for improved visualization.} 
\label{fig:qualitatives}
\end{figure}

Tab.~\ref{tab:cata_results} presents the results of the CATARACTS 2020 challenge. CAMI-SIAT uses additional phase annotations, while ARTG leverages tool labels for training. Our SR-Mamba outperforms other methods that only use step labels in terms of the average F1 score. Although ARTG gains better performance, using tool labels incurs a higher annotation cost in real clinical applications.

In Fig.~\ref{fig:qualitatives}, we display the color-coded ribbon of video 57 from the Cholec80 dataset. We also present results from TeCNO and Trans-SVNet. Our proposed method enjoys robustness against noisy observations (frames at \textcircled{1} and \textcircled{4}).

\subsection{Ablation Studies}
\label{sec:ablation}
We conduct ablation studies on the Cholec80 dataset to explore the effects of training sequence length and architectural variations. Tab.~\ref{tab:ablation_length} reveals that longer training sequences yield improvements in AC and JA, demonstrating the advantage of extended sequences.
\rui{Tab.~\ref{tab:ablation_anti} illustrates that incorporating anticipation loss into SR-Mamba results in approximately a 0.5 percentage point increase in both AC and JA.}

\begin{table}[t]
	\centering\footnotesize
	\caption{The results ($\%$) of different length number for training SR-Mamba on Cholec80 dataset~\cite{cholec80}.}
        \scalebox{1}{
        \input{Tables/ablation_length}
        }
\label{tab:ablation_length}
\end{table}

\begin{table}[t]
	\centering\footnotesize
	\caption{The results ($\%$) of without and with the anticipation task as an auxiliary on Cholec80 dataset~\cite{cholec80}.}
        \scalebox{1}{
        \input{Tables/ablation_anticipation}
        }
\label{tab:ablation_anti}
\end{table}

\begin{table}[!t]
	\centering\footnotesize
	\caption{The results ($\%$) of different architectures on Cholec80 dataset~\cite{cholec80}.}
        \scalebox{0.92}{
        \input{Tables/ablation_arch}
        }
\label{tab:ablation_arch}
\end{table}

Tab.~\ref{tab:ablation_arch} explores the impact of different architectures on surgical video recognition performance. In the first section, we investigate the sequence modeling capabilities of the bidirectional Mamba decoder (Bi-M) using a conventional two-step method. Specifically, we generate spatial features using a \rui{pretrained} ResNet50 or a ViT-B/16 backbone, followed by our proposed Mamba decoder for final output generation. The checkpoints for these two backbones are provided by their respective authors~\cite{jin2022trans,liu2023lovit}. The findings suggest that utilizing ResNet50 as the spatial feature extractor yields less favorable outcomes compared to the ViT-B/16, potentially due to overfitting issues. Conversely, ViT-B/16, with its advantage of \rui{temporally-rich training}~\cite{liu2023lovit}, not only showcases better performance but also affirms that the proposed bidirectional Mamba decoder can be effectively integrated into a two-step model alongside advanced spatial extractors to achieve comparable effectiveness.

The second part, shown in the third row of the table, examines the necessity of the bidirectional mechanism. Here, "Vanilla" refers to a Mamba block employing only a forward direction. This analysis highlights the crucial advantage of bidirectional modeling in surgical video recognition. The superior outcomes from the bidirectional method compared to the forward-only vanilla Mamba block underscore its importance in capturing comprehensive temporal relationships, thereby enhancing recognition accuracy.



%

\section{Conclusion}
In this study, we explored the application of SSMs for surgical video analysis, enjoying their linear scalability and GPU-friendly training attributes for long-sequence modeling. We introduced SR-Mamba, a simpler yet more accurate model that leverages a bidirectional Mamba decoder. SR-Mamba consistently achieves superior results on two publicly available surgical video analysis datasets, Cholec80 and the CATARACTS. This work pioneers the exploration of SSMs in the analysis of long-term videos, offering valuable insights for future advancements in this domain.

\rui{\subsubsection{Acknowledgements.}
This work is supported in part by the Shenzhen Portion of Shenzhen-Hong Kong Science and Technology Innovation Cooperation Zone under HZQB-KCZYB-20200089, the InnoHK of the Government of the Hong Kong Special Administrative Region via the Hong Kong Centre for Logistics Robotics, and the CUHK T Stone Robotics Institute.}



%
%

\bibliographystyle{splncs04}
\bibliography{reference.bib}





\end{document}

%% file: Tables/cholec80_results.tex
\begin{tabular}{cccccc}
\hline
\textbf{Method} & \textbf{Accuracy} & \textbf{Precision} & \textbf{Recall} & \textbf{Jaccard} & \textbf{\#Param} \\
\hline
MTRCNet-CL \cite{jin2020multi} & 89.2 $\pm$ 7.6 & 86.9 $\pm$ 4.3 & 88.0 $\pm$ 6.9 & - & 29.8M \\
OHFM \cite{yi2019hard} & 87.3 $\pm$ 5.7 & - & - & 67.0 $\pm$ 13.3 & 45.1M \\
TeCNO \cite{czempiel2020tecno} & 88.6 $\pm$ 7.8 & 86.5 $\pm$ 7.0 & 87.6 $\pm$ 6.7 & 75.1 $\pm$ 6.9 & 23.6M \\
Trans-SVNet \cite{jin2022trans} & 90.9 $\pm$ 5.8 & \textbf{91.4 $\pm$ 6.7} & 89.4 $\pm$ 6.5 & 79.7 $\pm$ 6.8 & 23.7M \\
Not End-to-End \cite{yi2022not} & 91.5 $\pm$ 7.1 & - & 86.8 $\pm$ 8.5 & 77.2 $\pm$ 11.2 & 23.8M\\
LoViT \cite{liu2023lovit} & 92.4 $\pm$ 6.3 & 89.9 $\pm$ 6.1 & 90.6 $\pm$ 4.4 & 81.2 $\pm$ 9.1 & - \\
SR-Mamba(Ours) & \textbf{92.6 $\pm$ 8.6} & 90.3 $\pm$ 5.2 & \textbf{90.6 $\pm$ 7.2} & \textbf{81.5 $\pm$ 8.6} & 21.3M \\
\hline
\end{tabular}

%% file: Tables/cata_results.tex
\begin{tabular}{ccccccc}
\hline
SK & Uniandes-BCV & Trans-SVNet \cite{jin2022trans} & CAMI-SIAT$\ast$ & ARTG$\ast$ & SR-Mamba (Ours)\\
\hline
0.8181 & 0.7827 & 0.8402 & 0.8242 & \textbf{0.8920} & \textbf{0.8538} \\
\hline
\end{tabular}

%% file: Tables/ablation_length.tex
\begin{tabular}{c|cccc}
\hline
Length (n) & \textbf{Accuracy} & \textbf{Precision} & \textbf{Recall} & \textbf{Jaccard} \\
\hline
512 & 90.7 $\pm$ 7.6 & 86.3 $\pm$ 7.5 & 88.5 $\pm$ 8.5 & 76.2 $\pm$ 11.7 \\
1024 & 90.9 $\pm$ 11.3 & 89.4 $\pm$ 5.7 & 90.4 $\pm$ 4.9 & 80.7 $\pm$ 7.1 \\
2048 & \textbf{92.6 $\pm$ 8.6} & \textbf{90.3 $\pm$ 5.2} & \textbf{90.6 $\pm$ 7.2} & \textbf{81.5 $\pm$ 8.6} \\
\hline
\end{tabular}

%% file: Tables/ablation_anticipation.tex
\begin{tabular}{c|cccc}
\hline
\multirow{2}*{\makecell[c]{Anticipation\\Loss}}  & 
\multirow{2}*{\textbf{Accuracy}} & \multirow{2}*{\textbf{Precision}} & \multirow{2}*{\textbf{Recall}} & \multirow{2}*{\textbf{Jaccard}} \\
& & & & \\
\hline
\xmark & 91.9 $\pm$ 7.6 & 89.8 $\pm$ 5.9 & 90.7 $\pm$ 4.1 & 81.1 $\pm$ 6.8 \\
\cmark  & \textbf{92.6 $\pm$ 8.6} & \textbf{90.3 $\pm$ 5.2} & \textbf{90.6 $\pm$ 7.2} & \textbf{81.5 $\pm$ 8.6} \\
\hline
\end{tabular}

%% file: Tables/ablation_arch.tex
\begin{tabular}{ccc|cccc}
\hline
\multirow{2}*{\makecell[c]{Feature\\Extraction}} & \multirow{2}*{\makecell[c]{Temporal\\Model}} & \multirow{2}*{\makecell[c]{Single-step\\Training}} & \multirow{2}*{\textbf{Accuracy}} & \multirow{2}*{\textbf{Precision}} & \multirow{2}*{\textbf{Recall}} & \multirow{2}*{\textbf{Jaccard}}\\
& & & & & & \\
\hline
ResNet50 & Bi-M & \xmark & 86.8 $\pm$ 7.3 & 85.5 $\pm$ 4.2 & 84.3 $\pm$ 8.5 & 71.1 $\pm$ 8.4 \\
ViT-B/16 & Bi-M & \xmark & 92.6 $\pm$ 6.5 & \textbf{90.5 $\pm$ 5.2} & 89.4 $\pm$ 8.1 & 80.9 $\pm$ 8.6 \\
ResNet34 & Vanilla & \cmark & 88.2 $\pm$ 10.8 & 88.1 $\pm$ 4.7& 85.7 $\pm$ 7.7& 75.2 $\pm$ 8.8\\
\hline
ResNet34 & Bi-M & \cmark & \textbf{92.6 $\pm$ 8.6} & 90.3 $\pm$ 5.2 & \textbf{90.6 $\pm$ 7.2} & \textbf{81.5 $\pm$ 8.6} \\
\hline
\end{tabular}